%% file: main.tex
\documentclass[conference]{IEEEtran}

\usepackage{placeins}
\usepackage{graphicx} 
\usepackage{amsmath}
\usepackage{booktabs}
\usepackage{dblfloatfix}

\input{preamble}
\usepackage[hidelinks]{hyperref}

\title{KitchenTwin: Semantically and Geometrically Grounded 3D Kitchen Digital Twins}

\author{
\IEEEauthorblockN{Quanyun Wu, Kyle Gao, Daniel Long, David A. Clausi, Jonathan Li, Yuhao Chen}
\IEEEauthorblockA{
University of Waterloo \\
Waterloo, ON, Canada
}
}

\begin{document}
\maketitle
\input{sec/0_abstract}

\input{sec/1_intro}

\input{sec/2_background_related_works}
\input{sec/3_Methodology}

\input{sec/4_Experiments}
\input{sec/5_conclusion}
{
    \small
    \bibliographystyle{IEEEtran}
    \bibliography{main}
}


\end{document}

%% file: sec/0_abstract.tex
\begin{abstract}

Embodied AI training and evaluation require object-centric digital twin environments with accurate metric geometry and semantic grounding. Recent transformer-based feedforward reconstruction methods can efficiently predict global point clouds from sparse monocular videos, yet these geometries suffer from inherent scale ambiguity and inconsistent coordinate conventions. This mismatch prevents the reliable fusion of these dimensionless point cloud predictions with locally reconstructed object meshes. We propose a novel scale-aware 3D fusion framework that registers visually grounded object meshes with transformer-predicted global point clouds to construct metrically consistent digital twins. Our method introduces a Vision-Language Model (VLM)-guided geometric anchor mechanism that resolves this fundamental coordinate mismatch by recovering an accurate real-world metric scale. To fuse these networks, we propose a geometry-aware registration pipeline that explicitly enforces physical plausibility through gravity-aligned vertical estimation, Manhattan-world structural constraints, and collision-free local refinement. Experiments on real indoor kitchen environments demonstrate improved cross-network object alignment and geometric consistency for downstream tasks, including multi-primitive fitting and metric measurement. We additionally introduce an open-source indoor digital twin dataset with metrically scaled scenes and semantically grounded and registered object-centric mesh annotations.
\end{abstract}

%% file: sec/1_intro.tex
\section{Introduction}
\label{sec:intro}
Embodied artificial intelligence and digital twin systems rely on realistic virtual environments for simulation and evaluation~\cite{mu2024robotwin, nasiriany2024robocasa}. Tasks such as autonomous navigation, object manipulation, and world-aware interaction require indoor scenes that are both geometrically consistent and semantically meaningful. In particular, robotic agents must reason about objects through representations that preserve global scene structure while maintaining object-level geometry and identity.
\begin{figure*}
    \centering
    \includegraphics[width=1\linewidth]{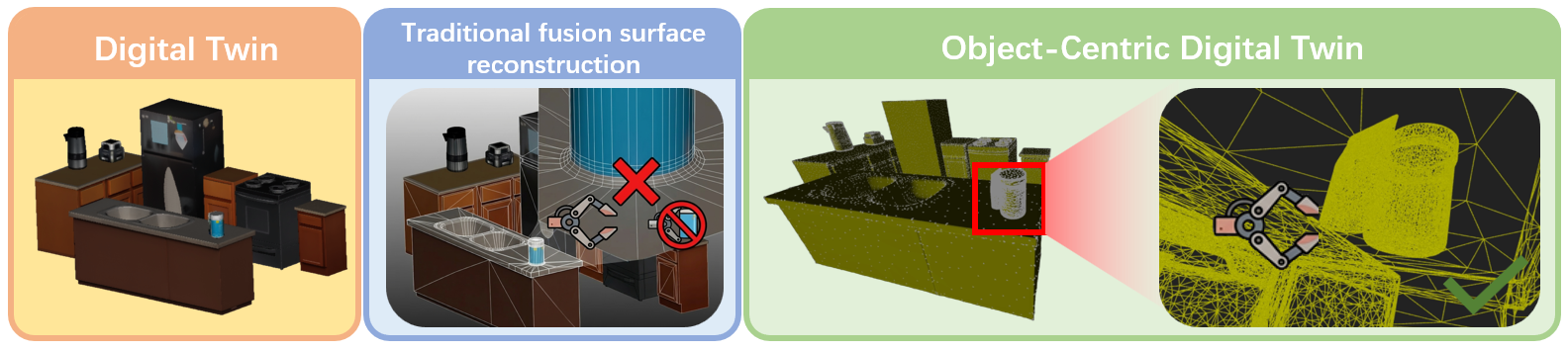}
    \caption{Comparison of scenarios. Left image: Output result of digital twin. Middle: Traditional fusion reconstruction method, in which objects (such as a bottle) share the geometric structure with the environment, thus unable to achieve physical independence. Right image: Our digital twin model based on objects, in which the objects are structurally independent and are physically operable meshes.}
    \label{fig:compare}
\end{figure*}
Large-scale indoor 3D datasets ~\cite{scannet, matterport3d, replica, hm3d} provide extensive coverage of real-world environments through reconstructed meshes, camera trajectories, and semantic annotations. However, as illustrated in Fig.~\ref{fig:compare}, these datasets primarily represent scenes as continuous, fused surface reconstructions. In this format, target objects are inextricably embedded within the global geometry—for example, a manipulable item sharing mesh vertices with the supporting counter. As a result, individual objects cannot be cleanly isolated, severely limiting their usefulness for embodied tasks such as item checking, delivery, and object-centric spatial reasoning. As a result, individual objects are often loosely defined or merged with surrounding surfaces, which limits their usefulness for object-centric semantic reasoning and manipulation. For embodied agents, objects must instead be represented as coherent meshes with well-defined geometry and identity.

For reliable embodied interaction, robotic systems operate with fixed kinematic constraints—such as gripper width and maximum reach—that must strictly correspond to the real-world metric scale of surrounding objects. However, modern 3D generative and feedforward reconstruction models inherently remove physical scale, operating instead within arbitrary, dimensionless coordinate spaces. Consequently, attempting to directly fuse these global scene reconstructions with locally generated object meshes introduces severe scale ambiguity, coordinate inconsistencies, and physically impossible mesh intersections.

Consequently, constructing reliable object-centric environments requires resolving three fundamental challenges: recovering the metric scale of reconstructed scenes, representing objects as complete and manipulable meshes, and enforcing consistent geometric alignment between object models and global scene geometry. Addressing these challenges is essential for generating physically plausible digital twins that support robust spatial reasoning and interaction for embodied agents.

\begin{figure*}[t]
    \centering
    \includegraphics[width=1\linewidth]{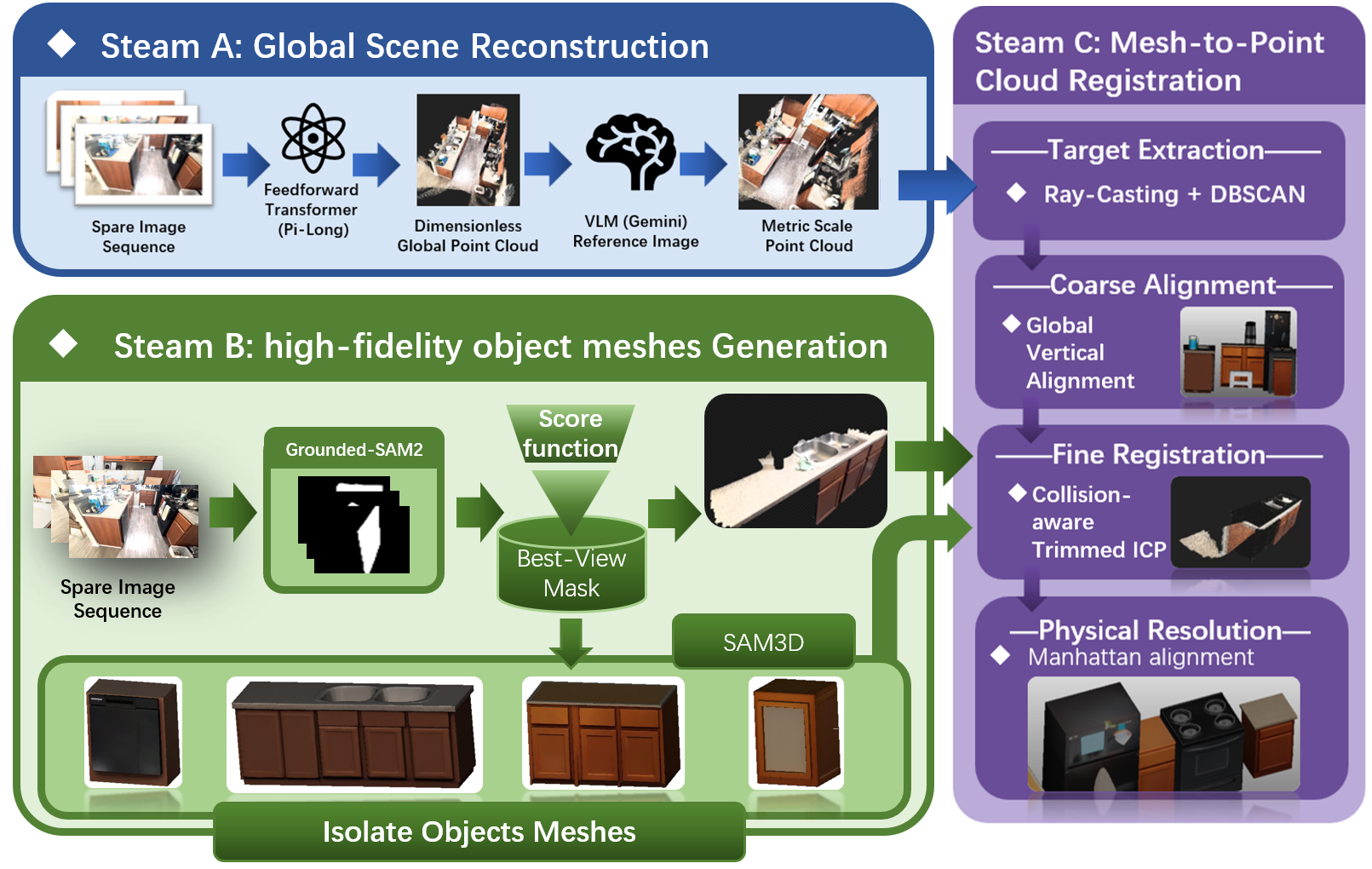}
    \caption{Our proposed pipeline for semantic and geometrically grounded object-centric scene reconstruction. Stream A (Top) reconstructs the global metric geometry via VLM-aided scale recovery. Stream B (Bottom) generates high-fidelity object meshes from optimal 2D views. Stage C (Right) fuses these streams through a geometrically grounded hierarchical registration process, enforcing physical plausibility and geometric constraints to produce a refined digital twin.}
    \label{fig:Pipeline}
\end{figure*}

To address these challenges, we propose a scale-aware 3D fusion framework that integrates local object meshes with globally reconstructed scenes in a consistent metric space. As illustrated in Fig.~\ref{fig:Pipeline}, our approach decomposes this complex fusion task into three synergistic streams, each designed to overcome a specific representational bottleneck:

\textbf{Stream A: Global Scene Reconstruction.} The primary challenge in utilizing feedforward transformer architectures is their inherent lack of absolute geometric scale. To overcome this, we first generate a dimensionless global point cloud, and then introduce a Vision-Language Model (VLM)-guided scale recovery module that explicitly transforms the scene into a geometrically accurate metric coordinate space $\mathcal{P}_{scaled}$.

\textbf{Stream B: Object Grounding and Mesh Generation.} Lifting objects from 2D images to 3D meshes requires mitigating severe occlusions and selecting optimal viewpoints to prevent geometric hallucinations. We solve this by employing an open-vocabulary tracking-and-selection mechanism to extract optimal, unoccluded multi-view masks, which are subsequently lifted into high-fidelity, structurally isolated object meshes $\mathcal{M}_i$.

\textbf{Stream C: Geometric Grounding.} The final and most critical bottleneck lies in fusing these isolated local meshes ($\mathcal{M}_i$) into the global metric space ($\mathcal{P}_{scaled}$) without introducing unphysical intersections or coordinate misalignments. To resolve this, we propose a geometrically constrained coarse-to-fine registration pipeline. By integrating gravity-aligned yaw hypotheses and Trimmed ICP (TrICP) with explicit structural constraints—such as Manhattan-world vertical alignment and collision resolution—our system effectively eliminates inter-object penetration, resulting in a physically plausible 3D digital twin.
\noindent
Our contributions are summarized as follows:
\begin{itemize}
    \item \textbf{Semantically and Geometrically Grounded 3D Reconstruction Framework:} We propose a 3D digital twin construction framework that effectively bridges the gap between different network architectures. It fuses global dimensionless surface point clouds with locally generated, complete object meshes into a coherent, manipulable environment.
    \item \textbf{Addressing Metric Ambiguity:} To ensure the virtual environment matches physical robotic dimensions, we introduce a Vision Language Model (VLM)-guided physical anchor mechanism. This successfully resolves scale ambiguity, transforming dimensionless surfaces into environments with accurate physical scale.
    \item \textbf{Geometry-Aware Registration:} To establish correct object relationships and ensure collision-free geometric alignment within the reconstructed scene. We design a cascade registration pipeline. Featuring world-vertical-aligned TrICP and collision resolution, our method enforces geometric grounding and eliminates inter-object penetrations.
    \item \textbf{Dataset Release:} We release KitchenTwin, an open-source digital twin dataset capturing a realistic North American kitchen environment. While large-scale datasets like EPIC-KITCHENS exist, our controlled capture setup is specifically designed for metrically accurate 3D evaluation. It enables comprehensive multi-view camera trajectories, precise semantic cataloging of items, and crucially, exact ground-truth scale verification through physical ruler measurements. The dataset provides metrically scaled scenes with semantic and geometric grounding, including RGB video sequences, 2D object masks, 3D point clouds, and explicitly registered 3D object meshes with per-object poses.
\end{itemize}

%% file: sec/2_background_related_works.tex
\section{Background and Related Works}

\subsection{Feedforward 3D Reconstruction Transformers}
Traditional 3D reconstruction relies on iterative multi-view optimization, which often suffers from scale drift and high computational costs. Recently, feedforward transformer architectures~\cite{sajjadi2022srt,transformer_3d_1, transformer_3d_2, zhang2024monst3r, yang2025fast3r} have emerged to directly and quickly estimate 3D geometry without explicit camera priors and per-scene training. Approaches like VGGT-Long~\cite{vggt_long_ref} further extend these capabilities to long sequences using efficient chunk-based processing.

\textbf{Pi-Long:} Pi-Long~\cite{pi_long_ref1} is a highly efficient feedforward transformer that reconstructs large-scale continuous point clouds and camera poses from monocular sequences in a single forward pass. While it provides robust global structures, its direct network predictions are inherently dimensionless. The lack of an absolute metric scale heavily restricts its direct application in downstream tasks with high geometric accuracy requirements.

\subsection{Semantically Conditioned 3D Generative Models}
Recent advancements in 3D content creation have shifted from complex optimization-based lifting to direct 3D native generation. Driven by vision-language alignment, modern generative models can directly synthesize dense 3D point clouds and high-fidelity meshes conditioned on semantic text prompts or single reference images, leveraging advances in 3D reconstruction. Works in the implicit 3D reconstruction era leveraging 2D generative models conditioned on text or images to generate mesh from learned implicit 3D models~\cite{jun2023shape,pooledreamfusion,melas2023realfusion,lin2023magic3d}.

\textbf{Direct 3D Object Generation:} Early works such \cite{nichol2022pointe, li2024craftsman} leverage explicit 3D point cloud representations and utilize a 3D native diffusion model to generate high-fidelity geometries with regular mesh topologies directly from single images. Similarly, GaussianAnything~\cite{yuan2025gaussiananything} leverages point-cloud flow matching to enable interactive, multi-modal 3D object generation, decoding inputs directly into dense point clouds and structured surfaces. 

Despite their remarkable ability to generate semantically accurate and geometrically detailed 3D objects, these native 3D generators share a critical limitation for embodied simulation: the generated meshes and point clouds inherently reside in arbitrary, isolated local coordinate systems (e.g., centered at the origin with unit bounding boxes). They entirely lack global spatial context, orientation priors, and real-world metric scale. Consequently, integrating these locally generated object meshes into a globally reconstructed physical scene remains a severe challenge, necessitating our proposed scale-aware registration framework.

\textbf{SAM3D \cite{sam3d_ref1}:} is a generative model that reconstructs 3D object geometry and texture from a single image and its corresponding mask that achieves state of the art geometric and semantic fidelity to the input image and masks. Although it yields precise geometry, the generated meshes inherently reside in isolated, arbitrary local coordinate systems (typically OpenGL format). They entirely lack global spatial context and physical scale, posing severe alignment challenges for scene integration.

\subsection{Digital Data Twins}
Digital twins serve as the fundamental infrastructure for training and evaluating embodied artificial intelligence~\cite{digital_twin_1}. For robotic agents to perform reliable spatial reasoning and physical manipulation, the simulated environments must be metrically precise and geometrically consistent. Currently, a critical gap exists between globally continuous but dimensionless reconstructed scenes and locally accurate but spatially isolated generative meshes. Our framework directly addresses this bottleneck by introducing a physics-aware registration pipeline to recover cross-network metric scale.

%% file: sec/3_Methodology.tex
\section{Methodology}
\label{sec:formatting} 
Our proposed framework constructs a scale-aware, object-centric 3D digital twin from a sparse monocular image sequence. The pipeline consists of three primary stages: (1) transformer-based global 3D point-cloud reconstruction; (2) semantic-object-based mesh generation via foundational and generative models; and (3) accurate mesh-to-point-cloud registration through collision-aware, vertical-aware, and occlusion-aware global-to-local alignment with metric grounding.

\subsection{Global Scene Reconstruction}
Traditional SLAM systems suffer from scale drift and require extensive multi-view optimization. Instead, we utilize a feedforward Transformer network, Pi-Long, to efficiently generate dense large-scale 3D point clouds from sequential images.

\textbf{Dimensionless Reconstruction.} Given an image sequence $I=\{I_1, I_2, \dots, I_N\}$, Pi-Long predicts depth maps, unscaled local point clouds, and relative camera poses in a single forward pass. For each chunk of frames, local point clouds are merged into a continuous dimensionless global point cloud $\mathcal{P}_{global}$ under the OpenCV coordinate convention (Right-Down-Forward). During this process, least-squares estimation on the network-normalized coordinates is used to recover robust camera intrinsics $K$.

\textbf{Physical Scale Recovery.} Since $\mathcal{P}_{global}$ is scale ambiguous, we introduce a physical anchor mechanism to recover metric scale. \textit{Gemini} configured with the \textit{google\_search\_retrieval} tool call is Vision-Language Model (VLM) with web search capabilities. It is queried with a reference image $I_k$ with a distinct anchor object to estimate its real-world width $w_{real}$ and depth $d_{real}$ together with a 2D bounding box $B_{2D}$.

Using the estimated intrinsics $K$ and the predicted camera-to-world pose $T_{c2w}$ for frame $k$, frustum culling is applied to crop $\mathcal{P}_{global}$ according to $B_{2D}$, producing the anchor point set $\mathcal{P}_{anchor}$. The maximum spatial extent of $\mathcal{P}_{anchor}$ yields the virtual width $w_{virtual}$. The global metric scale factor is computed as
\[
s=\frac{\sqrt{d_{real}^2+w_{real}^2}}{w_{virtual}} .
\]
A similarity transformation $T_{sim}\in Sim(3)$ is then constructed to scale all point coordinates and camera translations, producing the metric global point cloud $\mathcal{P}_{scaled}$.

\subsection{Semantic Object Grounding and Mesh Generation} We initialize open-vocabulary tracking on the first frame using \textit{Grounded-SAM-2}, which propagates spatial masks temporally across the sequence. To mitigate occlusions and noise, we implement an intra-frame merging algorithm based on Intersection over Union (IoU) and containment heuristics, aggregating highly overlapping masks of the same semantic class. Transient noise is further filtered by discarding tracklets appearing in fewer than $N_{min}$ frames.

Using \textit{SAM3D} to reconstruct a 3D mesh from a single 2D mask requires an optimal canonical viewpoint. We define a heuristic scoring function for each tracked mask $m$:$$Score(m) =$$ $$Area(m) \times W_{area} \times \left(1 - \text{Penalty}(m) \times \frac{W_{trunc}}{W_{area}}\right)$$where $Area(m)$ favors larger observations, and $\text{Penalty}(m)$ strictly penalizes masks truncated by image boundaries. Here, $\text{Penalty}(m) \in [0, 1]$ acts as a normalized truncation ratio that quantifies the degree to which an object’s mask intersects the image boundaries. It strictly prevents the system from selecting partially observed objects, which would otherwise cause severe geometric hallucinations during the 3D lifting process. The frame maximizing this score is selected as the canonical view. The optimal mask is then processed by a single-image-to-3D lifting model (\textit{SAM3D}) to generate a high-fidelity object mesh $\mathcal{M}_i$. Crucially, these local meshes inherently reside in isolated object-space OpenGL coordinate systems (Right-Up-Backward) and entirely lack global spatial context. This approach provides open-vocabulary semantically grounded 3D object meshes. 

 \begin{figure*}[t]
    \centering
    \includegraphics[width=0.8\textwidth]{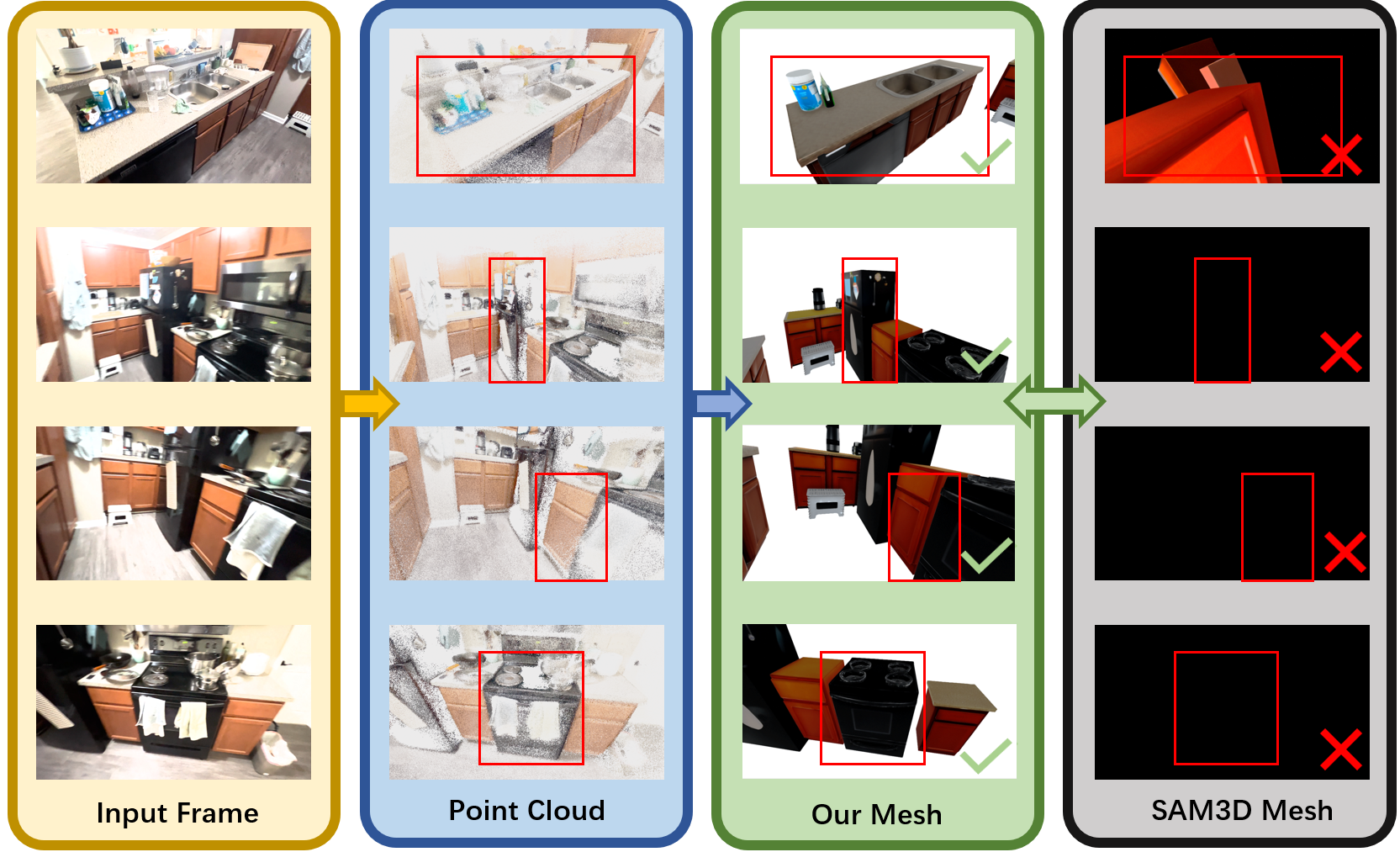}
    \caption{Visual progression of our 3D fusion framework compared to the baseline. From left to right: (1) The original \textbf{Input Frame} captured from the sequence. (2) The dense, unscaled \textbf{Point Cloud} reconstructed by Pi-Long. (3) \textbf{Our Mesh}, demonstrating that after scale recovery and geometries-aware registration, the lifted object meshes perfectly align within the physical 2D bounding box (red) when projected back to the camera view. (4) The naive \textbf{SAM3D Mesh} baseline, which fails to establish a coherent metric space, resulting in severe misalignment or absence within the ground truth bounding box when re-rendered.}
    \label{fig:nvss_masks}
\end{figure*}

\begin{table*}[t]
\centering
\caption{Quantitative comparison of 3D reconstruction semantic consistency. We report the Novel View Semantic Synthesis IoU (NVSS-IoU) for 12 individual objects and the overall mean IoU (mIoU). Our geometrically grounded alignment method significantly outperforms the baseline by establishing a coherent metric space.}
\label{tab:iou_comparison}
\resizebox{\textwidth}{!}{%
\begin{tabular}{l|cccccccccccc|c}
\toprule
\textbf{Method} & Bottle 1 & Fridge & Coffee M. & Cab. 1 & DishW. & Coffee & Cab. 2 & Stool & Cab. 3 & Oven & Bottle 2 & Cab. 4 & \textbf{mIoU} $\uparrow$ \\
\midrule
SAM3D & 0.0000 & 0.0000 & 0.0000 & 0.0000 & 0.2943 & 0.0000 & 0.0001 & 0.0000 & 0.0000 & 0.0000 & 0.0014 & 0.0000 & 0.0246 \\
\textbf{Ours} & \textbf{0.6321} & \textbf{0.6205} & \textbf{0.5751} & \textbf{0.1502} & \textbf{0.3979} & \textbf{0.7519} & \textbf{0.3079} & \textbf{0.6743} & \textbf{0.7798} & \textbf{0.5842} & \textbf{0.7283} & \textbf{0.6562} & \textbf{0.5715} \\
\bottomrule
\end{tabular}%
}
\end{table*}

\begin{table}[t]
\centering
\caption{Geometric registration metrics and mesh complexity. We report the TrICP Root Mean Square Error (RMSE), \textit{SAM3D}-to-World Scale Factor ($s$), and topological complexity (Vertices and Faces) of the generated meshes for all 12 tracked objects in the scene.}
\label{tab:registration_metrics}
\resizebox{\columnwidth}{!}{%
\begin{tabular}{l|c|c|cc}
\toprule
\textbf{Object} & \textbf{RMSE (m) $\downarrow$} & \textbf{Scale ($s$)} & \textbf{Vertices} & \textbf{Faces} \\
\midrule
Bottle 1        & 0.0165 & 0.24 & 3,970  & 7,028  \\
Bottle 2        & 0.0461 & 0.25 & 5,156  & 9,058  \\
Coffee Maker    & 0.0482 & 0.43 & 2,658  & 4,490  \\
Coffee          & 0.0910 & 0.43 & 6,016  & 10,336 \\
Cabinet 3       & 0.1637 & 0.69 & 4,165  & 7,054  \\
Step Stool      & 0.1699 & 0.45 & 27,366 & 35,710 \\
Stove/Oven      & 0.1780 & 0.85 & 37,162 & 56,524 \\
Cabinet 4       & 0.2072 & 0.84 & 4,377  & 7,652  \\
Cabinet 2       & 0.2440 & 2.10 & 5,618  & 8,594  \\
Fridge          & 0.3252 & 1.47 & 2,439  & 4,554  \\
Dishwasher      & 0.3264 & 0.77 & 10,303 & 17,975 \\
Cabinet 1       & 0.3657 & 1.25 & 8,762  & 14,270 \\
\bottomrule
\end{tabular}%
}
\end{table}

\begin{figure*}[t]
    \centering
    \includegraphics[width=0.45\textwidth]{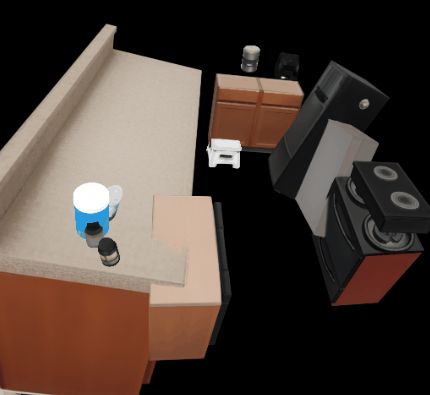}
    \includegraphics[width=0.52\textwidth]{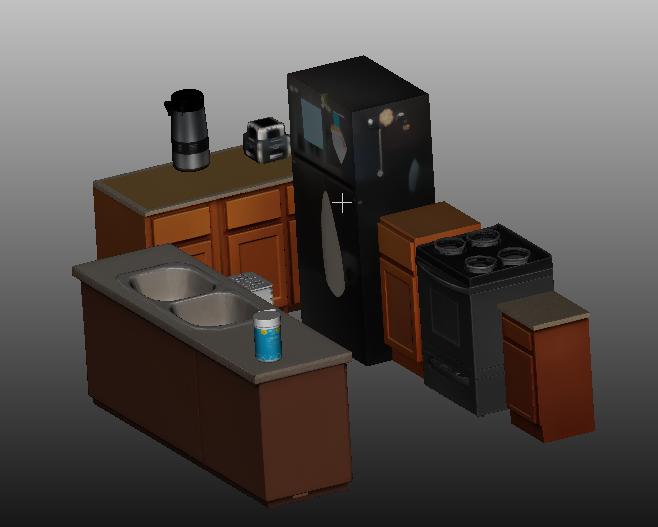}
    \caption{Qualitative comparison of the assembled 3D digital twin. \textbf{Left:} The baseline fails to establish a metric space, resulting in floating, unscaled, and overlapping artifacts. \textbf{Right:} With geometric grounding, our method produces a physically plausible, Manhattan-aligned, and tightly registered object-centric scene without subsurface penetration.}
    \label{fig:global_reconstruction}
\end{figure*}

 \subsection{Geometrically Grounded Mesh-to-Point Cloud Registration}

 The final stage embeds each isolated object mesh $\mathcal{M}_i$ into the metric global point cloud $\mathcal{P}_{scaled}$. 

 \textbf{Target Point Cloud Extraction.}
 For an object mesh $\mathcal{M}_i$ reconstructed from frame $k$, we use the corresponding 2D segmentation mask together with the camera pose $T_{c2w}^{k}$ and intrinsic matrix $K$ to cast rays into the global point cloud $\mathcal{P}_{scaled}$. Points intersecting the projected mask region are collected, and their depth distribution is analyzed. We then apply DBSCAN clustering to remove outliers and isolate the dominant cluster corresponding to the object. The resulting subset is denoted as the target point cloud $\mathcal{T}_i$.

 \textbf{Coarse Alignment.}
 The initial pose of $\mathcal{M}_i$ is arbitrary with respect to $\mathcal{P}_{scaled}$. We first apply the previously estimated scale factor obtained from comparing the spatial extents of $\mathcal{M}_i$ and $\mathcal{T}_i$. To reconcile coordinate conventions between the OpenGL mesh and the OpenCV point cloud, we estimate the global vertical axis by fitting a floor plane to $\mathcal{P}_{scaled}$ using RANSAC. The floor-plane normal defines the scene vertical direction. A rigid transformation is then applied to align the mesh up-axis with this vertical axis.

 Since the object's yaw rotation around the vertical axis remains ambiguous, we evaluate several orthogonal yaw hypotheses $\left(0, \frac{\pi}{2}, \pi, -\frac{\pi}{2}\right)$. For each hypothesis, the mesh is projected into the image plane using $T_{c2w}^{k}$ and $K$. The orientation that maximizes projection overlap with the object mask while maintaining front-facing consistency with the camera view is selected as the coarse initialization.

 \textbf{Fine Registration via TrICP.}
 Due to occlusions, $\mathcal{T}_i$ often represents only a partial surface of the object, whereas $\mathcal{M}_i$ is a complete mesh. Under such partial overlap conditions, standard ICP becomes unstable. We therefore employ Trimmed Iterative Closest Point (TrICP), which iteratively estimates the rigid transformation $T_{fine}$ by minimizing the alignment error over only the overlapping subset of correspondences:

 \[
 E(T_{fine}) = 
 \sum_{j=1}^{N_{overlap}}
 \left\|
 T_{fine} \, p_j^{\mathcal{M}} - q_j^{\mathcal{T}}
\right\|^2
 \]

 where $p_j^{\mathcal{M}} \in \mathcal{M}_i$ and $q_j^{\mathcal{T}} \in \mathcal{T}_i$. The optimization is constrained to planar translation and yaw rotation to maintain consistency with the scene floor and prevent unrealistic tilting.

 \textbf{Scene-Level Structural Consistency and Geometric Constraints}
 To enforce global layout consistency, we perform a final scene-level adjustment under a Manhattan-world assumption. The largest object in the scene (e.g., the kitchen counter) is selected as an anchor. The yaw angles of all other objects are snapped to orthogonal multiples of $\frac{\pi}{2}$ relative to this anchor to enforce axis-aligned layout structure.

 Finally, collision handling is performed in the horizontal plane using 2D bounding box intersection tests. Objects with large overlap ($>30\%$) with the anchor are classified as \emph{embedded} structures (e.g., ovens) and translated to lie flush with the anchor surface. Objects with smaller overlap are treated as \emph{freestanding} items and are displaced by a repulsion vector to eliminate intersection with neighboring objects.

%% file: sec/4_Experiments.tex
\section{Experiments}
\label{sec:experiments}

\subsection{Experimental Setup}
\textbf{Dataset:} We evaluate our framework on our novel real-world dataset featuring severe occlusions, diverse object scales, and complex physical layouts. The scene-level 3D reconstruction comprises a point cloud with 3354257 points.

\textbf{Baseline:} We compare our method against a \textit{SAM3D} baseline. In this setup, object meshes are generated via \textit{SAM3D} but are simply aggregated without metric scale recovery or physical collision resolution, leaving them in isolated, arbitrary local coordinate systems.

\textbf{Evaluation Metric (NVSS-IoU):} Direct 3D volumetric evaluation is challenging due to the lack of a perfect 3D ground truth. Therefore, we propose the \textbf{Novel View Semantic Synthesis IoU (NVSS-IoU)}. We re-render the reconstructed 3D scene back to the 2D camera plane using the original tracking poses. By applying the ground truth (GT) 2D bounding box as a physical prior, we prompt a foundation model (\textit{SAM3}) to extract the projected mask and compute the Intersection over Union (IoU) against the GT observation mask. NVSS-IoU strictly penalizes any micro-errors in 6-DoF spatial localization, metric scale, and geometric shape, serving as a highly rigorous metric for object-centric digital twins.

\subsection{Quantitative Results}
Table~\ref{tab:iou_comparison} presents the quantitative comparison of semantic consistency. The \textit{SAM3D} baseline completely collapses, yielding an abysmal mIoU of 0.0246. This catastrophic failure occurs because the generated meshes suffer from scale ambiguity and coordinate misalignment; when projected back to the camera plane, they completely miss the physical GT footprint (resulting in zero IoU for most objects). 

In contrast, our geometrically grounded alignment method achieves a substantial improvement, reaching an mIoU of \textbf{0.5715}. By resolving the OpenCV-to-OpenGL coordinate mismatch with a $Sim(3)$ transformation, applying TrICP, and enforcing collision constraints, our framework accurately anchors objects to their true metric coordinates, demonstrating robustness across diverse categories from small bottles to large cabinets.

Beyond semantic consistency, we evaluate geometric precision and topological fidelity across all tracked objects in the scene. Table~\ref{tab:registration_metrics} reports RMSE, recovered absolute metric scale ($s$), and mesh complexity (vertices and faces) for all 12 reconstructed objects. The consistently low RMSE values (e.g., 0.0165m for a small bottle, 0.1780m for a large oven) indicate that constrained TrICP with semantic collision handling reliably aligns high-fidelity meshes, ranging from thousands to tens of thousands of faces, even under heavy occlusion. The diverse scale factors (0.24–2.10) further validate the robustness of our physical anchor mechanism in mapping dimensionless geometries to accurate real-world metric scales.

\subsection{Qualitative Evaluation}

As shown in Figure~\ref{fig:global_reconstruction}, the baseline reconstruction produces a chaotic assembly of meshes due to the lack of spatial constraints. Conversely, our method reconstructs a geometrically coherent and geometrically grounded indoor environment. Furthermore, Figure~\ref{fig:nvss_masks} visualizes the step-by-step progression of our reconstruction and alignment process. When tracking the spatial projection of the objects, the unscaled point cloud provides a dense but dimensionless geometric foundation. By applying our geometry-aware registration, \textit{Our Mesh} achieves precise global-to-local spatial consistency; when re-rendered from the original camera poses, our aligned meshes project exactly into the ground truth physical bounding boxes (highlighted in red). In sharp contrast, the naive \textit{SAM3D Mesh} baseline completely fails to respect the global coordinate space, resulting in severe misalignments or empty projections within the target regions.


\subsection{Ablation Study}
To validate the efficacy of our core components, we analyze their individual contributions:
\begin{itemize}
    \item \textbf{w/o Physical Anchor:} Removing the API-driven scale recovery leaves the global point cloud dimensionless, causing TrICP to diverge and resulting in catastrophic size mismatches.
    \item \textbf{w/o TrICP:} Relying solely on standard ICP or coarse centroid alignment fails to handle the severe partial overlap between complete meshes and occluded target point clouds and results in complete registration failure in partially occluded scene areas.
    \item \textbf{w/o Geometrical Grounding:} Disabling collision resolution and ground leveling leads to severe mesh-to-scene surface collisions (e.g., embedded appliances protruding) and floating artifacts, which catastrophically degrade object registration.
\end{itemize}
Our NVSS-IoU metric directly measures the physical localization accuracy of reconstructed objects. Table \ref{tab:ablation_iou} shows that removing any major component leads to catastrophic mis-registration. Even in some cases with non-zero IoU, visual inspection reveals that overlap occurs by chance due to world and object coordinate initialization, while the object meshes remain unregistered. This confirms that all major components of our semantic and geometrically grounded registration are essential for functionally accurate digital twins.
\begin{table}[t]
\centering
\caption{Quantitative ablation of semantic and geometric consistency shows the Novel View Semantic Synthesis IoU (NVSS-IoU) for the full pipeline compared to three ablated variants. }
\label{tab:ablation_iou}
\resizebox{\columnwidth}{!}{%
\begin{tabular}{l|c|ccc}
\toprule
\textbf{Object} & \textbf{Ours (Full)} & \textbf{w/o metric Anchor} & \textbf{w/o TrICP} & \textbf{w/o Geom. Grounding} \\
\midrule
Bottle 1        & \textbf{0.6321} & 0.0015 & 0 & 0.4141 \\
Fridge          & \textbf{0.6205} & 0.2686 & 0 & 0.6048 \\
Coffee Maker    & \textbf{0.5751} & 0 & 0.0643 & 0.1393 \\
Cabinet 1       & \textbf{0.1502} & 0 & 0 & 0 \\
Dishwasher      & \textbf{0.3979} & 0 & 0 & 0  \\
Coffee          & \textbf{0.7519} & 0 & 0 & 0 \\
Cabinet 2       & \textbf{0.3079} & 0 & 0.0739 & 0 \\
Step Stool      & \textbf{0.6743} & 0.0062 & 0.0102 & 0.4605 \\
Cabinet 3       & \textbf{0.7798} & 0 & 0 & 0 \\
Stove/Oven      & \textbf{0.5842} & 0.2463 & 0 & 0 \\
Bottle 2        & \textbf{0.7283} & 0 & 0.5620 & 0 \\
Cabinet 4       & \textbf{0.6562} & 0 & 0.3874 & 0.5749 \\
\midrule
\textbf{mIoU ($\uparrow$)} & \textbf{0.5715} & 0.0436 & 0.0915 &0.1828 \\
\bottomrule
\end{tabular}%
}
\end{table}

%% file: sec/5_conclusion.tex
\section{Conclusion}
We presented a framework for constructing semantically and geometrically grounded digital twins from sparse monocular observations. Our approach resolves the disconnect between dimensionless global scene reconstructions and object-centric meshes defined in separate coordinate systems, enabling the construction of metrically consistent environments for embodied agents. By combining vision-language guided scale recovery with geometry-aware object registration, the system produces object-centric 3D scenes that preserve both semantic identity and geometric consistency within a shared physical frame. Experiments show improved object alignment and spatial consistency compared with naive mesh aggregation baselines. We also introduce the KitchenTwin dataset, which provides metrically scaled indoor scenes with reconstructed point clouds, registered object meshes, and object-centric annotations to support research in embodied AI and digital twin generation. 